\documentclass[10pt,twocolumn,letterpaper]{article}

\usepackage[pagenumbers]{cvpr} %

\usepackage{multirow}
\usepackage{url}            %
\usepackage{booktabs}       %
\usepackage{amsfonts}       %
\usepackage{nicefrac}       %
\usepackage{microtype}      %

\usepackage{times}
\usepackage{epsfig}
\usepackage{graphicx}
\usepackage{amsmath}
\usepackage{amssymb}
\usepackage{multirow}
\usepackage{colortbl}

\usepackage{microtype}
\usepackage{graphicx}
\usepackage{arydshln}
\usepackage{wrapfig}

\usepackage{subcaption}
\usepackage{pifont}

\definecolor{cvprblue}{rgb}{0.21,0.49,0.74}
\usepackage[pagebackref,breaklinks,colorlinks,allcolors=cvprblue]{hyperref}

\def\paperID{3243} %
\def\confName{CVPR}
\def\confYear{2026}

\title{Revisiting Real-Time Detection Transformer with Efficient Encoder Design}
\author{Jiannan Huang$^{1}$ \quad  Aditya Kane$^{1}$ \quad Fengzhe Zhou$^{1}$ \quad Yunchao Wei$^{2}$ \quad Humphrey Shi$^{1}$ \\ 
$^{1}$SHI Labs @ Georgia Tech $^{2}$Beijing Jiaotong University
}

\usepackage{xcolor}

\begin{document}
\maketitle
\begin{abstract}
Real-time object detection is crucial for real-world applications as it requires high accuracy with low latency.
While Detection Transformers (DETR) have demonstrated significant performance improvements, current real-time DETR models are challenging to reproduce from scratch due to excessive pre-training overheads on the backbone, constraining research advancements by hindering the exploration of novel backbone architectures.
In this paper, we want to show that by using general good design, it is possible to have \textbf{high performance} with \textbf{low pre-training cost}.
After a thorough study of the backbone architecture, we propose EfficientNAT at various scales, which incorporates modern efficient convolution and local attention mechanisms.
Moreover, we re-design the hybrid encoder with local attention, significantly enhancing both performance and inference speed.
Based on these advancements, we present Le-DETR (\textbf{L}ow-cost and \textbf{E}fficient \textbf{DE}tection \textbf{TR}ansformer), which achieves a new \textbf{SOTA} in real-time detection using only ImageNet1K and COCO2017 training datasets, saving about 80\% images in pre-training stage compared with previous methods.
We demonstrate that with well-designed, real-time DETR models can achieve strong performance without the need for complex and computationally expensive pretraining.
Extensive experiments show that Le-DETR-M/L/X achieves \textbf{52.9/54.3/55.1 mAP} on COCO Val2017 with \textbf{4.45/5.01/6.68 ms} on an RTX4090. It surpasses YOLOv12-L/X by \textbf{+0.6/-0.1 mAP} while achieving similar speed and \textbf{+20\%} speedup.  Compared with DEIM-D-FINE, Le-DETR-M achieves \textbf{+0.2 mAP} with slightly faster inference, and surpasses DEIM-D-FINE-L by \textbf{+0.4 mAP} with only \textbf{0.4 ms} additional latency. Code and weights will be open-sourced.
\begin{figure}
        \includegraphics[width=1.0\linewidth]{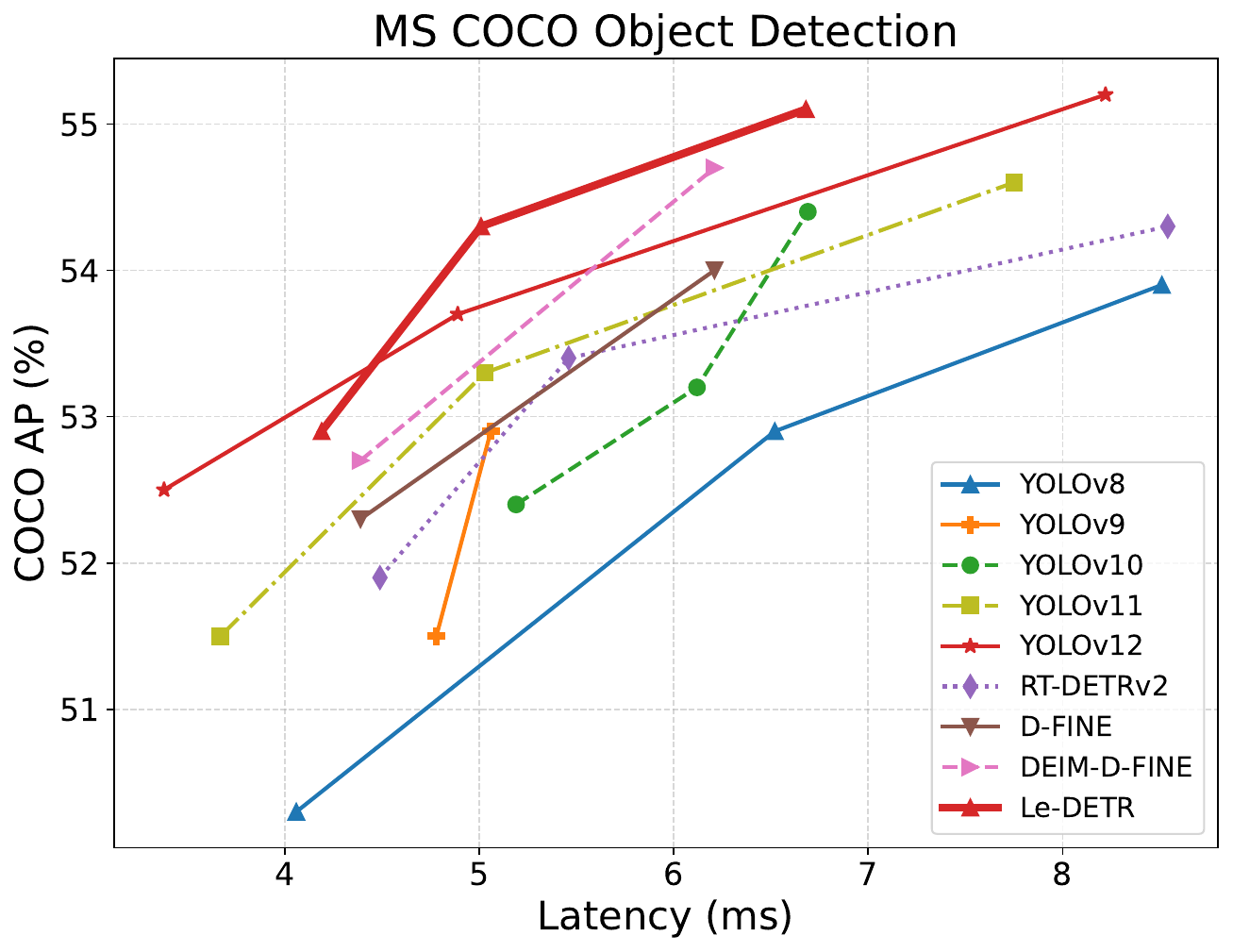}
        \caption{The performance of each model. The results show that our model outperforms the existing models and achieves a new SOTA for real-time detection models. These results are tested using vanilla PyTorch profiler on 1$\times$RTX4090.}
        \label{fig:performance}
        \vspace{-0.3cm}
    \end{figure}
\end{abstract}
    
\section{Introduction}
\label{sec:intro}

    The rapid advancements in deep learning have facilitated the development of highly effective real-time object detectors, most notably exemplified by the YOLO series~\cite{ultralytics_yolov5,li2022yolov6,wang2023yolov7,ultralytics_yolov8,wang2024yolov9,wang2024yolov10,ultralytics_yolov11,redmon2017yolo9000, wang2024gold, xu2022damo, MS-YOLO}. These models have significantly advanced the field by demonstrating the efficacy of Convolutional Neural Networks (CNNs) in real-time applications.
    In parallel, transformer-based detection models, such as the DETR~\cite{lv2024rt,wang2024rt,zhao2024detrs,peng2024dfine,zhang2022dino}, have shown powerful performance by eliminating complex post-processing steps, particularly Non-Maximum Suppression (NMS)~\cite{RCNN}, which has become a key bottleneck in accelerating inference. Several end-to-end models~\cite{wang2024yolov10, ultralytics_yolov11, lv2024rt, wang2024rt} have made significant progress in the field of real-time object detection.

    \begin{figure*}[tbp]
        \centering
        \includegraphics[width=1.0\textwidth]{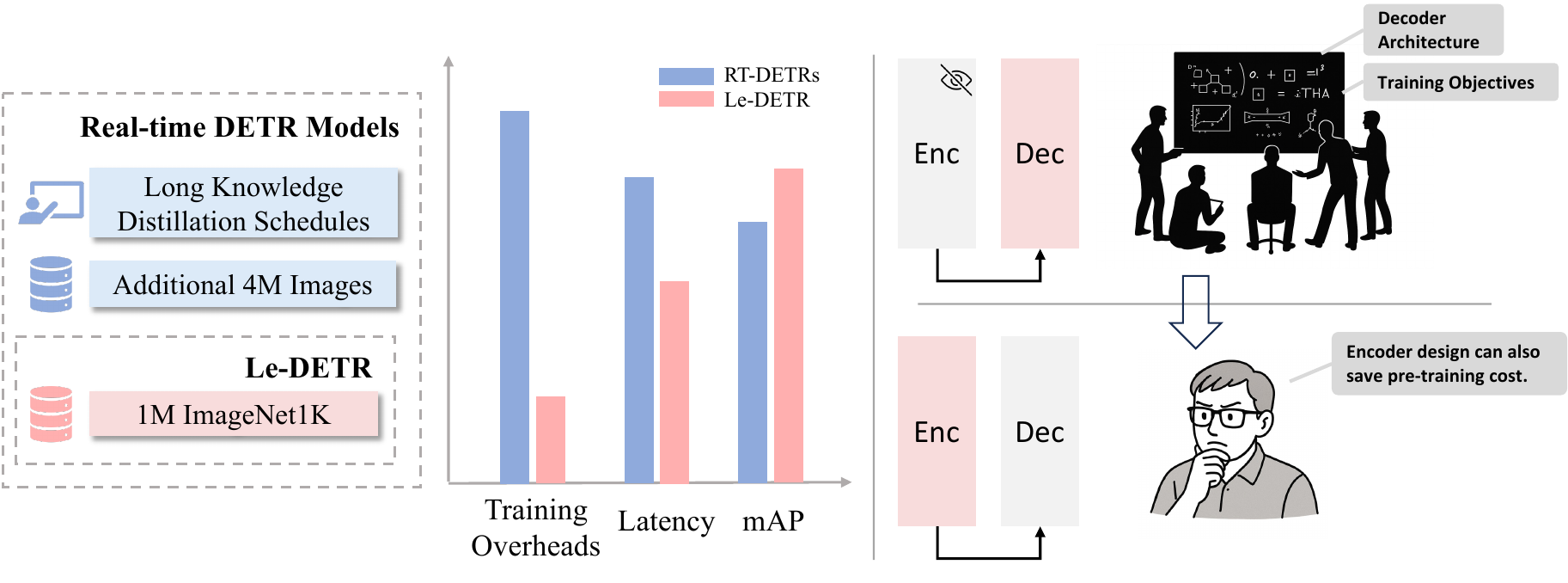}
        \caption{\emph{Left:} Comparison of Le-DETR and Real-Time DETR series in training overheads, While saving lots of training overheads offered by efficient encoder design, our model outperforms the previous SOTA model, D-FINE and DEIM-D-FINE.\emph{Right:} Unlike previous works such as DEIM and D-FINE mainly focus on decoder architecture and training objects, we focus on efficient encoder design.}
        \label{fig:overhead}
        \vspace{-0.3cm}
    \end{figure*}
    
    Although transformer-based real-time detectors demonstrate such impressive performance, they are challenging to reproduce from scratch and improve due to the excessive pertaining overheads in the backbone. 
    As illustrated in Fig.~\ref{fig:overhead}, real-time DETR models require an additional four million images~\cite{cui2021selfsupervisionsimpleeffectivenetwork}, approximately four times the size of the ImageNet 1K pretraining dataset. 
    Furthermore, these models depend on long knowledge distillation schedules, which require higher computational overheads~\cite {cui2021selfsupervisionsimpleeffectivenetwork} to complete the pretraining process.
    We ablate the key changes comparing with ResNet50 and ResNet50\_vd\_ssld~\cite{he2019bag,cui2021selfsupervisionsimpleeffectivenetwork}, which is used by RT-DETRv2-L as backbone:(1) Variants from ResNet50 to ResNet50-D; (2) Computationally expensive pre-training technologies. Results of Tab.~\ref{tab:pretrain}
    illustrates that a significant portion of their strong performance is derived from the substantial pre-training overhead (KD \&ATI). 
    This creates a significant barrier, effectively locking the community into specific, costly pre-training pipelines. It leads to a critical research question: Is this massive pre-training overhead a fundamental requirement for advanced performance, or is it merely a compensation for sub-optimal architectural designs in current models? 
    As shown in Fig.~\ref{fig:overhead}, though previous works like DEIM, and D-FINE focus majorly on decoder architecture and training objectives,
    we hypothesize that superior architectural design can break this dependency, achieving advanced performance using only ImageNet-1K pre-training.

    We found that the architectural modifications introduced by real-time DETR models primarily focus on CNN-style detection techniques, such as the FPN-PAN~\cite{liu2018path, lin2017feature} and the RepVGG-C3 block~\cite{ding2021repvgg}. However, there is a lack of emphasis on modern attention techniques, specifically local attention~\cite{hassani2023neighborhood, hassani2024faster, hassani2022dilated}, which limits models' ability to achieve good performance. This limitation arises from slow inference of attention mechanisms and insufficient exploitation of localized features.

\begin{table*}[tbp]
    \centering
    \resizebox{0.8\textwidth}{!}{
    \begin{tabular}{lccccc}
        \toprule[0.3mm]
        Model & KD \& ATI & Variant & Pretraining Image Nums &Latency(ms) & AP$^{val}$ \\
        \midrule
        \multirow{3}{*}{RT-DETRv2-L} & \ding{51} & \ding{51} & 5M &5.46 & 53.4 \\
        & \ding{55} & \ding{51} & 1M & 5.46 & 51.6 (\textcolor{red}{$\downarrow$1.8}) \\
        & \ding{55} & \ding{55} & 1M & 4.91 & 51.6 (\textcolor{red}{$\downarrow$1.8}) \\
        \midrule
        Le-DETR-L & \ding{55} & \ding{55} & 1M & 5.01 & 54.3 \\   
        \bottomrule[0.3mm]
    \end{tabular}}
    \caption{This table presents the performance of the RT-DETRv2 models without the use of knowledge distillation (KD) techniques and additional training images (ATI) for backbone pretraining. Variant means whether to use ResNet50-D~\cite{he2019bag}. 
    When compared to RT-DETRv2-L, our model achieves a \textbf{+0.9} mAP improvement. Under identical pre-training settings, the performance gap further widens, with our model surpassing RT-DETRv2-L by \textbf{+2.7} mAP.
    }
    \label{tab:pretrain}
    \vspace{-0.4cm}
\end{table*}

    In this paper, we demonstrate that the key to low-cost, high-performance real-time detection lies in the architectural design itself. By using general good design, it is possible to have advanced performance while having low pre-training overheads.
    To this end, we propose Le-DETR (\textbf{L}ow-Cost and \textbf{E}fficient \textbf{DE}tection \textbf{TR}ansformer). Our method is designed to significantly reduce the cost of training a real-time DETR model from scratch while having SOTA performance. Le-DETR leverages local attention techniques to improve the model performance and speed up the inference process. Within this backbone, we introduce a novel EfficientNAT module that integrates Neighborhood Attention with an MBConv feed-forward network (FFN) to enhance feature processing. Also, guided by previous work~\cite{cai2023efficientvit, radosavovic2020designing, hassani2023neighborhood}, we conducted a detailed backbone study to explore optimal backbone design choices.
    In addition, the hybrid encoder incorporates the Neighborhood Attention-based Improved Feature Inference (NAIFI) module, which is designed to strike an optimal balance between performance and latency by capitalizing on the advantages of Neighborhood Attention, also our models benefit from well-established implementations from the community.

    Extensive experiments demonstrate that our model exhibits remarkable performance, establishing a new SOTA in real-time object detection, with saving 3M, about 80\% images for pre-traing the backbone, showing that the excessive pre-training overheads are not necessary for advanced performance. Specifically, Le-DETR-M/L/X achieves 52.9/54.3/55.1 mAP and 4.73/5.01/6.68 ms on an RTX4090. As shown in Fig.~\ref{fig:performance}, compared with previous SOTA in yolo series, it outpeforms YOLOv12-L/X by 0.6, -0.1 mAP by having -0.02\%, 20\% speedup, respectively. Compared to previous SOTA in DETR series, it outpeforms DEIM-D-FINE-M by 0.2 mAP with slightly speedup, LE-DETR-X out perform DEIM-D-FINE-L by 0.4 mAP with only 0.4 ms slower. Extensive ablation experiments validate the effectiveness and non-redundancy of our proposed improvements, and experiments on the backbone scale design investigated how to scale up and scale down the EfficientNAT efficiently.

    \noindent
    To be summarized, our contributions are as follows:
    \begin{itemize}
        \item We identify a commonly overlooked limitation of real-time DETR models: their substantial pre-training overheads on backbone, which have hindered further innovation within the research community. 
        We show that even not specially design for low pre-training data, it is possible to have SOTA performance with only pre-training on ImageNet1K, reducing $\sim$80\% training images. It can boost reproducibility and architectural innovation in this field.
        \item We propose a novel backbone for real-time detection that substantially reduces the computational costs associated with enhancing DETR models after a detailed backbone study.
        \item We show that local attention is effective in real-time detection. Using local attention, we redesign the encoder of the model to enhance the performance of Le-DETR.
    \end{itemize}
    \vspace{-0.1cm}

\section{Related Works}
\label{sec:rel}

    \subsection{Real-Time Detection}
    Real-time detection aims at localizing and classifying objects in given images with low latency. Due to the high speed and powerful performance of model inference, real-time detection models are crucial in real-world applications. Extensive works~\cite{redmon2016you, redmon2017yolo9000, redmon2018yolov3, bochkovskiy2020yolov4,ultralytics_yolov5, ultralytics_yolov8, lv2024rt, wang2024yolov10, zhao2024detrs, li2022yolov6, wang2023yolov7, wang2024yolov9, xu2022damo, wang2024gold, wang2024rt, jiao2024collaborativevisiontextrepresentationoptimizing} have explored the field of real-time detection. Specifically, the YOLO series models show powerful performance. The models of YOLOv1~\cite{redmon2016you}, YOLOv2~\cite{redmon2017yolo9000}, and YOLOv3~\cite{redmon2018yolov3} form the paradigm of modern real-time detectors: backbone, neck, and head.
    YOLOv5~\cite{ultralytics_yolov5} mainly increases model performance by introducing CSPNet~\cite{wang2020cspnet}.
    YOLOv8~\cite{ultralytics_yolov8} incorporated transformer-based layers and introduced C2F blocks, enhancing the extraction of features for improved precision in complex environments. YOLOv10~\cite{wang2024yolov10} introduced a two-stage supervision to build a NMS-free~\cite{felzenszwalb2009object} detection model.
    Beyond conv-based YOLO series models, transformer-based real-time detection models show powerful performance. RT-DETR~\cite{zhao2024detrs} outperforms YOLOv8~\cite{ultralytics_yolov8} mainly by replacing the original Deformable DETR~\cite{zhu2020deformable} decoder in DINO~\cite{zhang2022dino, carion2020end} with the widely used FPN-PAN~\cite{lin2017feature, liu2018path} network and a novel RepVGG~\cite{ding2021repvgg} block. RTDETR-v2~\cite{lv2024rt} further improved performance by setting up a set of bag-of-freebies.
    Although real-time DETR models have made great progress, they always focus on integrating advanced technologies in object detection, while lacking consideration of advanced efficient transformer technologies. In addition, they highly rely on a backbone that is trained with large amounts of data, the complex training strategies, and expensive training costs hinder the community from further improving models.
    Our work focuses on using modern attention technologies to design a new model to further improve the reproducibility and performance of transformer-based real-time detection models.

    \begin{figure*}[tbp]
        \centering
        \includegraphics[width=1.0\textwidth]{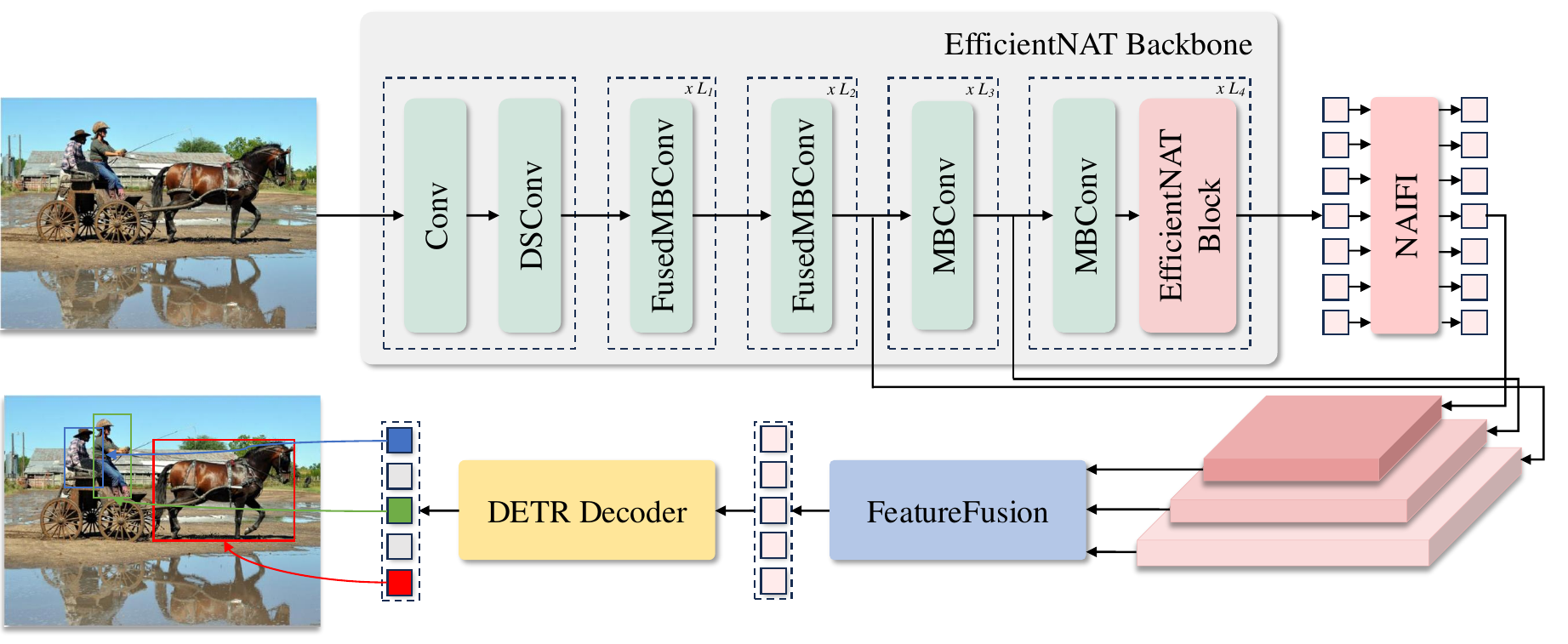}
        \caption{The proposed Le-DETR is structured in three distinct stages: backbone, encoder, and decoder. Each first block of stages in this backbone serves as the downsampling block. In this figure, we also illustrate the encoder, which incorporates both the NAIFI and Feature Fusion components.}
        \label{fig:model}
        \vspace{-0.3cm}
    \end{figure*}
    
    \begin{figure}
        \centering
        \includegraphics[width=\linewidth]{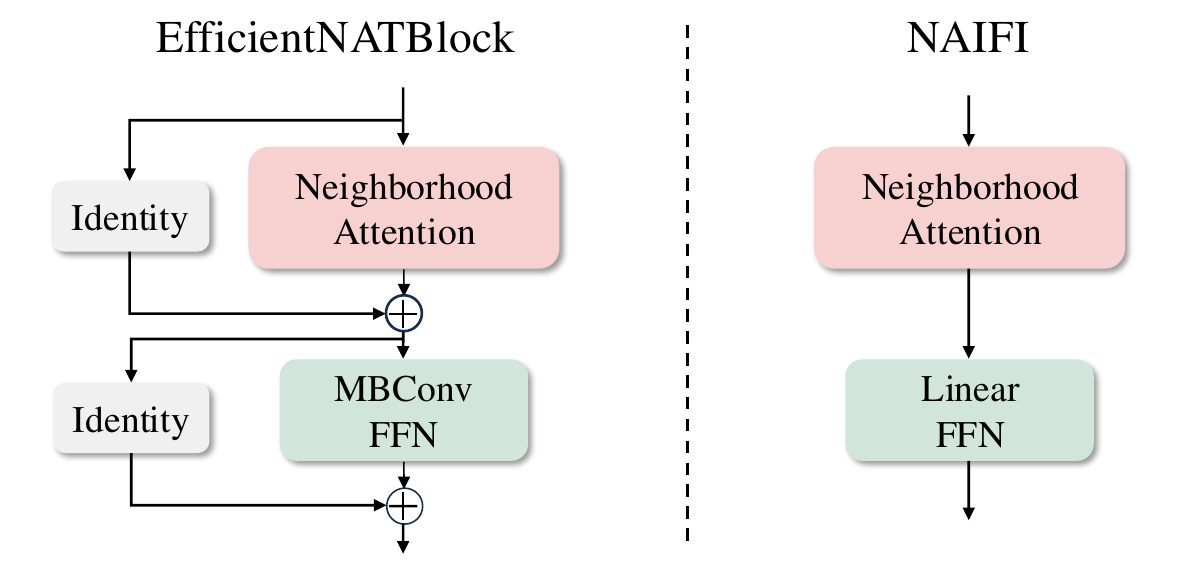}
        \caption{Overview of EfficientNATBlock and Neighborhood Attention-based Improved Feature Inference(NAIFI).}
        \label{fig:com}
        \vspace{-0.3cm}
    \end{figure}

    \subsection{Efficient Attention}
    Attention~\cite{vaswani2017attention} has proven to be a highly effective mechanism for capturing long-range dependencies and enhancing feature representation in models. In addition, it helps transformer-based models show a powerful scaling capability.
    Previous work has also demonstrated the power of attentional mechanisms in computer vision.
    The Vision Transformer (ViT)~\cite{dosovitskiy2020image} introduced a fully attention-based architecture, showing that transformers could outperform traditional convolutional networks in image classification by capturing global dependencies. Building on this, the Swin Transformer~\cite{liu2021swin} utilized a hierarchical structure with shifted windows, enabling attention to be computed efficiently in a local context while still capturing global information.
    Furthermore, the Detection Transformer (DETR)~\cite{zhao2024detrs, zhang2022dino, zhu2020deformable} revolutionized object detection by formulating it as a set prediction problem, eliminating the need for traditional anchors and post-processing steps through the use of transformer-based attention, thereby simplifying the detection pipeline.
    However, self-attention is slow because its computational complexity scales quadratically with the input size, requiring $O(n^2)$ operations to compute attention weights for all pairs of input tokens. Specifically, a BMM-style implementation of self-attention -- one having back-to-back matrix multiplications with softmax in between -- is slowed down due to an excessive number of slow read/writes to global memory.
    Much work~\cite{hassani2023neighborhood, hassani2022dilated, hassani2024faster, xFormers2022, dao2022flashattention, dao2023flashattention, shah2024flashattention} has explored how to accelerate self-attention-based raining and inference.
    Fused attention~\cite{dao2022flashattention, dao2023flashattention, shah2024flashattention} accelerates self-attention by fusing the matrix multiplications and softmax into a single kernel, thereby reducing global memory read/writes while being functionally equivalent and reducing the quadratic complexity without sacrificing accuracy. Specifically, Neighborhood Attention ~\cite{hassani2023neighborhood, hassani2022dilated, hassani2024faster,hassani2025generalizedneighborhoodattentionmultidimensional} reduces computational overhead by focusing attention on a fixed local window, limiting the scope to nearby elements while maintaining spatial context.

\section{Method}
\label{sec:method}
    \subsection{Preliminary}

        \paragraph{Neighborhood Attention} Neighborhood Attention (NA) \cite{hassani2023neighborhood} is an approach within the attention framework, emphasizing localized feature aggregation by concentrating attention on a small, spatial neighborhood around each query position. Unlike standard attention mechanisms, which consider all possible context positions, NA reduces theoretical computational complexity by limiting the scope of attention to relevant, proximal areas. This method leverages spatial correlations more efficiently and improves both computational efficiency and the preservation of local structural information. Given an input $X \in \mathbb{R}^{n \times d}$, where $X$ is a matrix whose rows are $d$-dimensional token vectors. Linear projections are applied to obtain $\mathbf{Q}$, $\mathbf{K}$, and $\mathbf{V}$ from $X$. With these projections and relative positional biases $B(i, j)$, the attention weights for the $i$-th input with a neighborhood size $k$, denoted as $A_i^k$, are defined, and its $k$ nearest neighborhood are:
        \begin{equation}
            \mathbf{A}_i^k =
            \begin{bmatrix}
            Q_i K_{\rho_1(i)}^T + B(i, \rho_1(i)) \\
            Q_i K_{\rho_2(i)}^T + B(i, \rho_2(i)) \\
            \vdots \\
            Q_i K_{\rho_k(i)}^T + B(i, \rho_k(i))
            \end{bmatrix}
        \end{equation}
    
        And $ \rho_j(i) $ denotes $i$’s $j$-th nearest neighbor, The attention weight is computed as the dot product between the query of the $i$-th input and its $k$ nearest neighboring keys. The neighboring values, $\mathbf{V}_i^k$, are then defined as a matrix whose rows are the value projections of the $k$ nearest neighbors of the $i$-th input:
        \begin{equation}
            \mathbf{V}_i^k = \left[ V_{\rho_1(i)}^T \quad V_{\rho_2(i)}^T \quad \cdots \quad V_{\rho_k(i)}^T \right]^T.
        \end{equation}

        Neighborhood Attention for the $i$-th token with neighborhood size $k$ is then defined as:
        
        \begin{equation}
            \text{NA}_k(i) = \text{softmax}\left( \frac{\mathbf{A}_i^k}{\sqrt{d}} \right) \mathbf{V}_i^k.
        \end{equation}
  
    \subsection{Analysis of Pretraining on Backbone}

        Compared to real-time DETR models and the DETR~\cite{carion2020end}, a significant distinction is the reliance on extensively pre-trained backbones in addition to modifications made to the DETR encoder~\cite{lv2024rt, cui2021selfsupervisionsimpleeffectivenetwork}. Specifically, these detection models only employ PResNet50/101\_vd\_ssld~\cite{cui2021selfsupervisionsimpleeffectivenetwork,he2019bag}\footnote{\href{https://paddleclas.readthedocs.io/en/latest/models/ResNet_and_vd_en.html}{https://paddleclas.readthedocs.io/en/latest/models/ResNet\_and\_vd\_en.html}}, or PP-HGNet\_v2\_ssld~\cite{cui2021selfsupervisionsimpleeffectivenetwork}\footnote{\href{https://github.com/PaddlePaddle/PaddleClas/blob/release/2.6/docs/en/models/PP-HGNet_en.md}{https://github.com/PaddlePaddle/PaddleClas/blob/release/2.6/docs/en/models/PP-HGNet\_en.md}} instead of ResNet50 as their backbone. 
        Such backbone~\cite{cui2021selfsupervisionsimpleeffectivenetwork} is initially pre-trained on a dataset comprising four million filtered unlabeled images, followed by fine-tuning on ImageNet1K, which includes about one million images.
        This approach substantially increases the computational and data pre-training overheads for training a real-time detection model from scratch, as shown in Fig.~\ref{fig:overhead}. Although the pre-trained weights derived from these complex pipelines enable high performance, they also pose challenges for researchers aiming to improve model efficiency and performance, as the reliance on such large-scale pre-training pipelines complicates replication, experimentation, and innovation. 
        Also, the 4M unlabeled filtered images are not open-sourced, which hinders the fair comparison and further improvement by the community.
        Consequently, the current real-time DETR models are largely confined to using either PP-HGNetv2 or PResNet as the backbone~\cite{lv2024rt, wang2024rt, peng2024dfine, zhao2024detrs}, thereby limiting the exploration of novel backbone designs and hindering potential advancements in the field.
        
        In contrast, the widely adopted and cost-effective training approach within the community involves pre-training solely on ImageNet1K, which contains approximately one million labeled images—only one-fifth of the data used by previous real-time DETR models, as shown in Fig.~\ref{fig:overhead}. 
        In this paper, we want to answer one question: Is it possible to have SOTA performance while having low pre-training cost? In the following section, we show that through careful backbone design and study, and comprehensive re-design of the overall model architecture, it is possible.

    \begin{table*}[t]
    \centering
    \resizebox*{0.97\linewidth}{!}{
    \begin{tabular}{lccccccccc}
        \toprule[0.5mm]
        Models & Params (M) & GFLOPs & Latency(ms) & AP$^{val}$ & AP$^{val}_{50}$ & AP$^{val}_{75}$ & AP$^{val}_{S}$ & AP$^{val}_{M}$ & AP$^{val}_{L}$ \\
        \midrule
        \midrule
        \multicolumn{10}{c}{\textit{\textbf{YOLO Series Models}}} \\\noalign{\vskip 6pt}
        YOLOv5-M~\cite{ultralytics_yolov5} & 25.1 & 64.2 & \underline{3.57}  & 49.1 & 66.0 & 53.8 & 31.2 & 54.2 & 65.4\\
        YOLOv5-L~\cite{ultralytics_yolov5} & 46.5 & 109.1 & 5.57 & 49.0 & 67.3 & -  & - & - & -\\
        YOLOv5-X~\cite{ultralytics_yolov5} & 86.7 & 205.7 & 8.55 & 50.7 & 68.9 & -  & - & - & - \\
        \hdashline\noalign{\vskip 2pt}
        YOLOv8-M~\cite{ultralytics_yolov8} & 25.9 & 78.9 & 4.06 & 50.3  & 67.3 & 54.8 & 32.3 & 55.9 & 66.5 \\
        YOLOv8-L~\cite{ultralytics_yolov8} & 43.7 & 165.2 & 6.52 & 52.9 & 69.8 & 57.7 & 35.5 & 58.5 & 69.8 \\
        YOLOv8-X~\cite{ultralytics_yolov8} & 68.2 & 257.8 & 8.51 & 53.9 & 71.1 & 58.9 & 36.0 & 59.4 & 70.9 \\
        \hdashline\noalign{\vskip 2pt}
        YOLOv9-M~\cite{wang2024yolov9} & 20.1	& 76.8 & 4.78 & 51.5 & 68.4 & 54.0 & 33.8 & \underline{57.2} & 67.3 \\
        YOLOv9-C~\cite{wang2024yolov9} & 25.5	& 102.8 & 5.06 & 52.9 & 69.8 & 57.7 & 35.6 & 58.2 & 69.0 \\
        \hdashline\noalign{\vskip 2pt}
        YOLOv10-B~\cite{wang2024yolov10} & 15.4 & 59.1  & 5.19 & 52.4 & 69.5 & 57.1 & 35.0 & 57.7 & 68.3 \\
        YOLOv10-L~\cite{wang2024yolov10} & 24.4 & 120.3 & 6.12 &  53.2 & 70.1 & 58.1 & 35.8 & 58.5 & 69.4 \\
        YOLOv10-X~\cite{wang2024yolov10} & 29.5 & 160.4 & \underline{6.69} &  54.4 & 71.3 & 59.3 & 37.0 & \underline{59.8} & 70.9 \\
        \hdashline\noalign{\vskip 2pt}
        YOLO11-M~\cite{ultralytics_yolov11} & 20.1 & 68.0 & 3.67 & 51.5 & 68.5 & 55.7 & 33.4 & 57.1 & 67.9 \\
        YOLO11-L~\cite{ultralytics_yolov11} & 25.3  & 86.9 & 5.03 & 53.3 & 70.1 & 58.2 & 35.6 & 59.1 & 69.2 \\
        YOLO11-X~\cite{ultralytics_yolov11} & 56.9 & 194.9 & 7.75 & 54.6 & 71.6 & 59.5 & 37.7 & 59.7 & 70.2 \\
        \hdashline\noalign{\vskip 2pt}
        YOLOv12-M~\cite{tian2025yolov12} & 20.2 & 67.5 & \textbf{3.38} & 52.5 & 69.6 & 57.1 & \textbf{35.7} & \textbf{58.2} & 68.8 \\
        YOLOv12-L~\cite{tian2025yolov12} & 26.4 & 88.9 & \textbf{4.89} & 53.7 & 70.7 & 58.5 & \underline{36.9} & \underline{59.5} & 69.9 \\
        YOLOv12-X~\cite{tian2025yolov12} & 59.1 & 199.0 & 8.22 & \textbf{55.2} & 72.0 & \textbf{60.2} & \textbf{39.6} & \textbf{60.7} & 70.9 \\
        \hdashline\noalign{\vskip 2pt}
        YOLOv13-L~\cite{tian2025yolov12} & 27.6 & 88.4 & 7.33 & 53.4 & 70.9 & 58.1 &  - & - & - \\
        YOLOv13-X~\cite{tian2025yolov12} & 64.0 & 199.2 & 11.89 & 54.8 & 72.0 & \underline{59.8} & -  & - & - \\
        \midrule
        \multicolumn{10}{c}{\textit{\textbf{Transformer-based Models}}}\\\noalign{\vskip 6pt}
        RT-DETRv2-M$^{*}$~\cite{lv2024rt} & 36.6 & 100.3 & 4.49 & 51.9 & 69.9 & 56.5 & 33.5 & 56.8 & 69.2 \\
        RT-DETRv2-L~\cite{lv2024rt} & 42.9 & 137.3 & 5.46 & 53.4 & 71.6 & 57.4 & 36.0 & 57.9 & 70.8 \\
        RT-DETRv2-X~\cite{lv2024rt} & 76.5 & 260.0 & 8.54 & 54.3 & \textbf{72.8} & 58.8 & 35.8 & 58.8 & \textbf{72.1} \\
        \hdashline\noalign{\vskip 2pt}
        RT-DETRv3-M$^{*}$~\cite{wang2024rt} & 36.6 & 100.3 & 4.49 & 51.7 & - & - & - & - & - \\
        RT-DETRv3-L~\cite{wang2024rt} & 42.9 & 137.3 & 5.46 & 53.4 & - & - & - & - & - \\
        RT-DETRv3-X~\cite{wang2024rt} & 76.5 & 260.0 & 8.54 & 54.6 & - & - & - & - & - \\
        \hdashline\noalign{\vskip 2pt}
        D-FINE-M~\cite{peng2024dfine} & 19.2 & 56.6 & 4.39 & 52.3 & 69.8 & 56.4 & 33.2 & 56.5 & \textbf{70.2} \\
        D-FINE-L~\cite{peng2024dfine} & 30.7 & 91.0 & 6.21 & 54.0 & 71.6 & 58.4 & 36.5 & 58.0 & \textbf{71.9} \\
        \hdashline\noalign{\vskip 2pt}
        DEIM-D-FINE-M~\cite{huang2024deim} & 19.2 & 56.6 & 4.39 & \underline{52.7} & \textbf{70.0} & \underline{57.3} & 35.3 & 56.7 & \underline{69.5} \\
        DEIM-D-FINE-L~\cite{huang2024deim} & 30.7 & 91.0 & 6.21 & \textbf{54.7} & \textbf{72.4} & \textbf{59.4} & 36.9 & \textbf{59.6} & \underline{71.8} \\
        \hdashline\noalign{\vskip 2pt}
        Le-DETR-M (Ours) & 31.4 & 114.1 & 4.45 & \textbf{52.9} & \underline{70.0} & \textbf{57.4} & \underline{35.4} & 56.9 & 68.6 \\
        Le-DETR-L (Ours) & 41.5 & 124.3 & \underline{5.01} & \underline{54.3} & \underline{71.7} & \underline{58.9} & \textbf{37.1} & 58.7 & 71.4 \\
        Le-DETR-X (Ours) & 44.9 & 196.9 & \textbf{6.68} & \underline{55.1} & \underline{72.5} & 59.8 & \underline{38.1} & 59.6 & \underline{71.4} \\
        \bottomrule[0.5mm]
    \end{tabular}}
    \caption{Comparison of our proposed Le-DETR and previous SOTA object detection models on COCO Val 2017. Experiments show that our model Le-DETR-X outperforms the previous SOTA YOLOs and DEIM.}
    \label{tab:core_exp}
    \vspace{-0.4cm}
\end{table*}

    \subsection{Overview of Le-DETR}

    Le-DETR follows the standard design paradigm of real-time DETR models, comprising an efficient backbone, an encoder, and a decoder. As shown in Fig.~\ref{fig:model}, it provides a comprehensive overview of the proposed Le-DETR. In Section~\ref{sec:backbone}, we detail the architecture of our efficient backbone, termed EfficientNAT, which is carefully designed. Out of its performance, we can get comparable performance with lower pre-training cost. 
    Furthermore, in Section~\ref{sec:encoder&decoder}, we describe how local attention mechanisms are employed in both the encoder and decoder. This dual emphasis on efficiency and performance positions Le-DETR as a leading approach in real-time detection models.

    \subsection{EfficientNAT Backbone}
    \label{sec:backbone}

    \noindent\textbf{Architecture Design} As shown in Tab.~\ref{tab:pretrain}, the existing backbone is not sufficient to have good performance with low pre-training overheads. As a consequence, reducing pre-training overhead while achieving improved performance requires the introduction of a well-designed backbone capable of both low-latency inference and the extraction of robust multi-scale image features for the encoder.
    We begin designing our backbone based on EfficientViT~\cite{cai2023efficientvit, zhang2024efficientvit}, a series of SOTA backbones that strike an effective balance between latency and performance on ImageNet1K.
    To further enhance inference speed and model performance, we investigate the use of neighborhood attention, a variant of local attention, in our proposed backbone to extract more robust features in the final stage. This leads to the development of our proposed EfficientNAT. The EfficientNAT architecture achieves these objectives by incorporating several efficient convolutional operations in the first three stages, culminating in an attention module in the fourth and final stage. Specifically, we employ depthwise separable convolutions (DSConv) in the network's stem to project input images into a feature space. Consistent with common practices~\cite{tan2021efficientnetv2, cai2023efficientvit, zhang2024efficientvit}, we utilize Fused Mobile Convolution in the first two stages and standard Mobile Convolution in deeper stages, resulting in progressively smaller feature maps with an increasing number of channels. The first block in each stage functions as a downsampling layer, halving the spatial dimensions while doubling the number of channels. In the final stage of our architecture, a Mobile Convolution (MBConv) downsampling module, initiates the process, followed by a series of EfficientNAT blocks. The EfficientNATBlock module combines Neighborhood Attention with an MBConv, serving as a Feed-Forward Network (FFN) to function as a highly efficient yet powerful feature extractor.

    \noindent\textbf{Backbone Architecture Design}
    To ensure the designed backbone achieves both high throughput and strong performance, model architecture—particularly the number of blocks per stage—is crucial. To efficiently scale the backbone up or down for variants of SOTA detection models, we conduct a detailed design study to determine the optimal architecture.
    Specifically, in line with common practices in the community, there are three prominent patterns for scaling the model up or down: A balanced distribution~\cite{cai2023efficientvit}: the third and fourth stages have the same number of blocks, notated as $\mathrm{P}_\mathrm{A}$, a late-stage heavy distribution~\cite{radosavovic2020designing}: the fourth stage has more blocks than the third, notated as $\mathrm{P}_\mathrm{B}$, and an early-stage heavy distribution~\cite{radosavovic2020designing, hassani2023neighborhood,hassani2022dilated, he2015deepresiduallearningimage}: the third stage has more blocks than the fourth, notated as $\mathrm{P}_\mathrm{C}$. To the best of our knowledge, there is no established principle to accurately deduce which pattern is best for each model scale without any empirical experiments. Therefore, we conduct multiple experiments to identify the best pattern for each scale. Detailed experimental design principle and quantitative experiment results are shown in Section~\ref{sec:bbdesgin}.
    Based on the experimental results, we find that when scaling EfficientNAT to a large scale (X), using more blocks in the third stage ($\mathrm{P}_\mathrm{C}$) yields better performance, while using the same number of blocks in the third and fourth stages ($\mathrm{P}_\mathrm{A}$) performs better for the smaller scale (L) model. Additionally, we apply this conclusion when designing the even smaller (M) model.

    \subsection{NAIFI \& DETR Decoder}
    \label{sec:encoder&decoder}

    We want to explore how the hybrid encoder can benefit from modern attention mechanisms, specifically, from local attention.
    To this end, we introduce a redesigned attention mechanism, termed Neighborhood Attention-based Improved Feature Inference (NAIFI). NAIFI is a single-layer neighborhood attention transformer designed to optimize feature representation. This module not only improves feature extraction quality but also accelerates processing speed by using a relatively small kernel size, thereby facilitating rapid inference. Additionally, within the DINO~\cite{zhang2022dino} framework, a prediction head is assigned to each decoding layer during training, enabling the reduction of decoder layers used during inference, which further optimizes the model's efficiency. And we use Flash Attention~\cite{dao2023flashattention} in the decoder to speed up the self-attention inference process.

\begin{table*}[tbp]
    \centering
    \resizebox{\textwidth}{!}{
    \begin{tabular}{lccccccccc}
        \toprule[0.3mm]
        Models & Params (M) & GFLOPs & Latency(ms) & AP$^{val}$ & AP$^{val}_{50}$ & AP$^{val}_{75}$ & AP$^{val}_{S}$ & AP$^{val}_{M}$ & AP$^{val}_{L}$ \\
        \midrule
        Le-DETR \emph{w/} ResNet50\_vd\_ssld & 43.8 & 128.3 & 5.80 & 53.6 & \underline{71.4} & 58.0 & 35.5 & 58.3 & 70.9\\
        Le-DETR \emph{w/} EfficientViT & 64.7 & 135.7 & 6.60 & 53.8 & \underline{71.5} & \underline{58.6} & \underline{36.8} & 58.1 & \underline{71.3} \\
        Le-DETR \emph{w/o} NAIFI & 41.5 & 124.3 & \underline{5.18} & \underline{54.1} & 71.2 & \textbf{58.9} & 36.3 & \underline{58.6} & 70.4\\
        \midrule
        Le-DETR (Ours Full Method) & 41.5 & \textbf{124.3} & \textbf{5.01} & \textbf{54.3} & \textbf{71.7} & \textbf{58.9} & \textbf{37.1} & \textbf{58.7} & \textbf{71.4} \\
        \bottomrule[0.3mm]
    \end{tabular}}
    \caption{Ablation experiment of components in our proposed Le-DETR. These experiments show the effectiveness of each module by offering both model performance improvement or speedup model inference while keeping the same performance.}
    \label{tab:abl}
    \vspace{-0.4cm}
\end{table*}

 \begin{table}
    \centering
    \resizebox{\linewidth}{!}{
    \begin{tabular}{lccccc}
        \toprule
        Model Name & Params(M) & GFLOPs & Latency(ms) &  AP$^{val}$ & AP$^{val}_{50}$ \\
        \midrule
        Le-DETR-M$^{4}$ & 31.4 & 114.1 & 4.19 & 52.5 & 69.3 \\
        Le-DETR-M$^{5}$ & 32.7 & 115.0 & 4.45 & 52.9 & 70.0 \\
        Le-DETR-M$^{6}$ & 34.0 & 115.6 & 4.73 & 52.9 & 70.0\\
        \hdashline\noalign{\vskip 2pt}
        Le-DETR-L$^{4}$ & 38.8 & 122.6 & 4.53 & 53.7 & 70.9 \\
        Le-DETR-L$^{5}$ & 40.1 & 123.5 & 4.79 & 54.2 & 71.6 \\
        Le-DETR-L$^{6}$ & 41.5 & 124.3 & 5.01 & 54.3 & 71.7 \\
        \hdashline\noalign{\vskip 2pt}
        Le-DETR-X$^{4}$ & 42.3 & 195.2 & 6.10 & 54.6 & 71.8 \\
        Le-DETR-X$^{5}$ & 43.6 & 196.0 & 6.42 & 54.9 & 72.3 \\
        Le-DETR-X$^{6}$ & 44.9 & 196.9 & 6.68 & 55.1 & 72.5\\
        \bottomrule
    \end{tabular}}
    \caption{Results of the ablation experiments using different number of layers in the model decoder, we use six decoder layers for training, with corner labels corresponding to different inference layers.}
    \vspace{-0.4cm}
    \label{tab:layers}
\end{table}

\section{Experiments}
\label{sec:exp}
    \subsection{Experimental Setup}

    \noindent \textbf{Implementation Details} We follow the commonly DETR training process, employing supervised pre-training on ImageNet1K and then training the model on COCO train2017 dataset. For evaluation, we employ the COCO dataset. Model performance is assessed using the traditional COCO mean Average Precision (mAP) metric, which provides a comprehensive measure of accuracy across various object classes.
    To further elucidate the model's capabilities, we report COCO mAP metrics at multiple scales, thereby evaluating performance under different conditions.
    To accelerate inference, we adopt a 5-layer decoder architecture in Le-DETR-M. Within this decoder, we incorporate two training enhancements from D-FINE~\cite{peng2024dfine}: Fine-grained Distribution Refinement (FDR) and Global Optimal Localization Self-Distillation (GO-LSD).
    Also, wo use the Matchability-Aware Loss (MAL), which is proposed by DEIM~\cite{huang2024deim}, as our training loss.
    Detailed hyper-parameters  are in Appendix.
    \vspace{0.1cm}

    \noindent \textbf{Dataset \& Metric} To evaluate the performance of Le-DETR, we report detailed metrics including the total number of parameters and GFLOPs, which serve as indicators of the model's computational complexity.
    For latency testing, we measure inference delay using the PyTorch profiler in FP16 precision on an RTX 4090 GPU.
    In accordance with established practices in real-time detection, we utilize an input shape of (640, 640) for both GFLOPs and latency benchmarks to ensure consistency with typical operational scenarios. Specifically, we employ the \texttt{calc-flops} repo~\cite{calculate_flops_pytorch} to compute the GFLOPs and parameter counts of our model.
    \subsection{Quantitative Experiments}
    
                \begin{table*}[tbp]
    \centering
    \begin{minipage}{0.53\textwidth}
        \resizebox{\textwidth}{!}{
        \begin{tabular}{clcccr}
            \toprule
            \multicolumn{2}{c}{Model Name} & Block Number & Top1 Acc & Top5 Acc & Throughput \\
            \midrule
            \multirow{4}{*}{L} 
             & P$_\mathrm{A}\text{-}1$ \textcolor{green}{\ding{51}} & (1, 1, 4, 4)  & 81.994 & 95.432 & 1798.75 \\
             & P$_\mathrm{A}\text{-}2$ & (1, 1, 6, 6)  & 82.135 & 95.302 & 1607.68 \\
             & P$_\mathrm{B}$   & (1, 1, 4, 8)  & 82.046 & 94.972 & 1524.13 \\
             & P$_\mathrm{C}$   & (1, 1, 10, 2) & 81.946 & 95.494 & 1278.94 \\
            \midrule
            \multirow{9}{*}{X} 
             & P$_\mathrm{A}\text{-}1$   & (2, 2, 8, 8)   & 82.300 & 95.224 & 1247.82 \\
             & P$_\mathrm{A}\text{-}2$   & (2, 2, 10, 10)   & 82.238 & 95.086 & 1166.25 \\
             & P$_\mathrm{A}\text{-}3$   & (2, 2, 12, 12)   & 82.083 & 94.906 & 1059.28 \\
             & P$_\mathrm{B}\text{-}1$ & (2, 2, 8, 10)  & 82.178 & 95.008 & 1211.05 \\
             & P$_\mathrm{B}\text{-}2$ & (2, 2, 4, 12)  & 82.152 & 94.840 & 1183.47 \\
             & P$_\mathrm{B}\text{-}3$ & (2, 2, 2, 15)  & 81.744 & 94.744 & 1162.34 \\
             & P$_\mathrm{C}\text{-}1$ & (2, 2, 15, 2)  & 82.638 & 95.768 & 1058.14 \\
             & P$_\mathrm{C}\text{-}2$\ \textcolor{green}{\ding{51}} & (2, 7, 15, 2)  & 82.902 & 95.960 & 1006.83 \\
             & P$_\mathrm{C}\text{-}3$ & (2, 7, 18, 2)  & 82.750 & 95.890 & 950.25 \\
            \bottomrule
        \end{tabular}}
    \end{minipage}
    \hfill
    \begin{minipage}{0.45\textwidth}
        \centering
        \includegraphics[width=\textwidth]{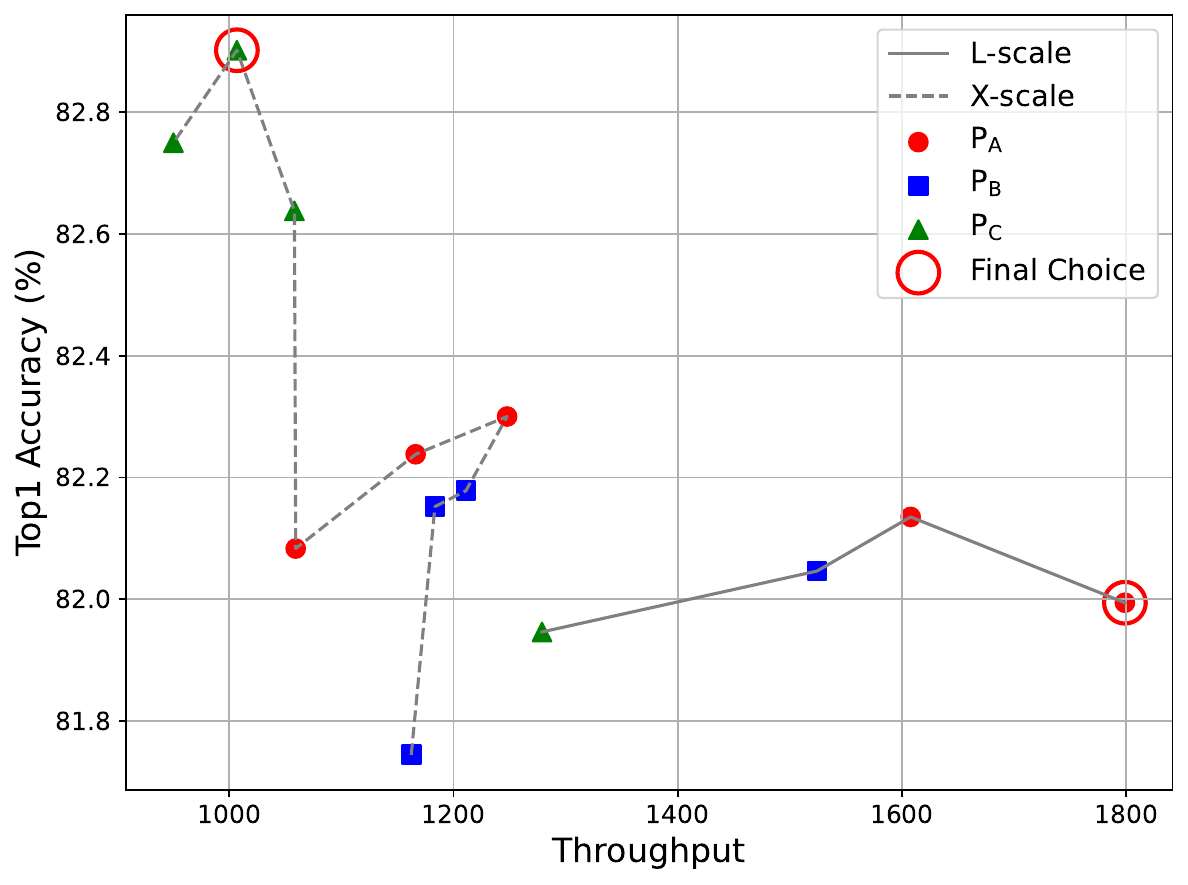}
    \end{minipage}
    
    \caption{We conducted experiments to explore different patterns on the backbone. As described in Section~\ref{sec:backbone}, we designed our backbone scale study using three distinct patterns. These are referred to as P$_\mathrm{A}$, P$_\mathrm{B}$, and P$_\mathrm{C}$, corresponding to balanced distribution, late-stage heavy distribution and an early-stage heavy distribution patterns, respectively. The results demonstrate that, for the L-scale, P$_\mathrm{A}$ yields the best performance, whereas P$_\mathrm{C}$ for the X-scale. The images clearly show the preference of different patterns at different scales. }
    \vspace{-0.5cm}

    \label{tab:backbone}
\end{table*}

        In this section, we present a comparative analysis of our model against SOTA real-time detectors. Specifically, we evaluate our approach in comparison to the well-established YOLO series models as well as end-to-end transformer-based real-time detectors (RT-DETRs). Tab.~\ref{tab:core_exp} summarizes the experimental results. Consistent with prior work, our analysis focuses on models at the M, L and X scales.
        Notably, Le-DETR-M achieves a mean Average Precision (mAP) of 52.4 with a processing time of 4.45 ms when tested on an RTX 4090, Le-DETR-L achieves 53.6 mAP and 5.01 ms, and Le-DETR-X attains a mAP of 54.7 with a processing time of 6.68 ms on the same hardware.
        Compared with YOLO11-M, Le-DETR-M is 1.4 mAP better, and for YOLO11-L, Le-DETR-L has 1.0 mAP better performance and comparable latency,
        Le-DETR-X is 0.5mAP better than YOLO11-X, and it is 14\% faster than it.
        Compared with YOLOv12, Le-DETR-M is 0.4 mAP better than YOLOv12-M, for YOLOv12-L, Le-DETR-L has 0.6mAP better performance with similar latency, and Le-DETR-X is 20\% fatser than YOLOv12-X with comparable performance.
        In comparison to SOTA real-time DETR models, Le-DETR-L has 0.7 mAP better performance than RT-DETRv2/v3-L while being 9\%  faster, with Le-DETR-M is 1.0/1.2 mAP better.
        Similarly, Le-DETR-X outperforms RT-DETRv2-X by 0.8 mAP, RT-DETRv3 by 0.5 mAP and is 22\% faster speed than RT-DETR-v2/v3-X.
        For D-FINE models, Le-DETR-M outperforms it by 0.6 mAP and with only 0.3 ms slower.
        Le-DETR-L achieves a 0.3 mAP improvement over D-FINE-L and 20\% faster. 
        Compared with DEIM-D-FINE-M, Le-DETR-M is 0.2 mAP better than it and faster, Le-DETR-X is 0.4 mAP better thank DEIM-D-FINE-L with only 4\% slower.
        Moreover, Le-DETR offers a significant advantage in terms of reproducibility, as it can be trained from scratch with substantially reduced pre-training overhead compared to existing real-time DETR models due to the efficient encoder design.

    \subsection{Ablation Experiments}

        \subsubsection{Ablation experiments on individual components}

        \label{sec:abl_components}

        \textbf{EfficientNAT Backbone} We conducted two experiments to assess the effectiveness of modifications to the backbone. The first row of Tab.~\ref{tab:abl} presents the results for the proposed overall backbone. By replacing EfficientNAT with ResNet50\_vd\_ssld, which is used in RT-DETR-v1/2/3, the latency increases from 5.01 ms to 5.80 ms, while the mAP decreases from 54.3 to 53.6. This performance reduction is primarily due to the decline in mAP$_{S}$ and mAP$_{M}$. Additionally, we compare our proposed EfficientNAT with the EfficientViT-L1. The second row of Tab.~\ref{tab:abl} shows the corresponding results. Using the EfficientViT-L1 as the backbone increases latency from 5.01 ms to 6.60 ms, while mAP decreases slightly from 54.3 to 53.8. These experiments demonstrate the necessity for proposing a new backbone and the performance of EfficientNAT.
        \vspace{0.1cm}

        \noindent \textbf{NAIFI} We replace the proposed NAIFI with the original AIFI, as introduced in RT-DETR, within the hybrid encoder. The third row of Tab.~\ref{tab:abl} presents the results. When using the original AIFI, mAP decreases from 54.3 to 54.1, while latency increases from 5.01 ms to 5.18 ms. The experimental results indicate that: (1) Using local attention instead of self-attention reduces latency, thereby making the model inference process faster; and (2) Local attention in the encoder improves model performance. These findings demonstrate the efficiency of our proposed NAIFI. We will discuss the selection of kernel type and size in the Appendix.

        \subsubsection{Ablation Experiments on Backbone Design}

        \label{sec:bbdesgin}
        In this section, we detail the backbone scale design process. As described in Section~\ref{sec:backbone}, we categorize our experiments into three patterns: $\mathrm{P}_{\mathrm{A}}$, balanced distribution; 
        $\mathrm{P}_{\mathrm{B}}$, late-stage heavy distribution;
        and $\mathrm{P}_{\mathrm{C}}$, early-stage heavy distribution These architecture schemes are elucidated in Tab.~\ref{tab:backbone}. In each experiment, we aim to maintain a constant number of blocks in the first two stages, while exploring different combinations to identify the optimal pattern for each scale.
        We started from three patterns corresponding to each design principle -- $\mathrm{P}_{\mathrm{A}}\text{-}1$, $\mathrm{P}_{\mathrm{B}}\text{-}1$ and $\mathrm{P}_{\mathrm{C}}\text{-}1$, and then modified the same to maintain a decent throughput-accuracy balance.
        For the L scale, we observed $\mathrm{P}_{\mathrm{A}}\text{-}1$ to have high throughput, and acceptable accuracy. Owing to this, we experimented with $\mathrm{P}_{\mathrm{A}}\text{-}2$, which performed the best but shows a relatively low throughput compared with $\mathrm{P}_{\mathrm{A}}\text{-}1$. Furthermore, we conducted experiments using them on COCO. Experiment results show that they have the same mAP, while the model using $\mathrm{P}_{\mathrm{A}}\text{-}1$ is faster (4.53 ms vs 4.96 ms).
        Thus, we chose $\mathrm{P}_{\mathrm{A}}\text{-}1$ as our final architecture. For the X scale, we observed $\mathrm{P}_{\mathrm{C}}\text{-}1$ to outperform other configurations and thus scaled it to $\mathrm{P}_{\mathrm{C}}\text{-}2$, which was chosen as our final configuration. However, we found that scaling it to  $\mathrm{P}_{\mathrm{C}}\text{-}3$ resulted in a drop in accuracy. Finally, we choose $\mathrm{P}_{\mathrm{C}}\text{-}2$ for X-scale.
        The figure in Tab.~\ref{tab:backbone} clearly visualizes the pattern selection.

        \subsubsection{Ablation Experiments on Decoder Layers}
        To evaluate the model's performance under different numbers of inference decoder layers, we conducted inference decoder layer ablation experiments. All models are first trained using six decoder layers. Tab.~\ref{tab:layers} presents the results, demonstrating that Le-DETR can flexibly adjust the trade-off between accuracy and efficiency in practical applications. Also, it shows that using 5 layers in decoder, Le-DETR-M has 0.26ms speedup without harming the performance.

\section{Conclusion}

This paper introduces Le-DETR, an efficient real-time object detection model that addresses key limitations of transformer-based detection architectures. Le-DETR reduces plenty of pre-training overheads while having great performance in real-time detection.
Our experiments demonstrate that Le-DETR has a series of models with SOTA performance, with Le-DETR-M achieving 52.9 mAP and 4.45 ms, Le-DETR-L surpassing previous models by achieving 54.3 mAP with a 5.01 ms latency, and Le-DETR-X achieves 55.1 mAP with 6.68 ms latency.
We discuss the limitation in the appendix. Future research could further optimize transformer-based models to minimize or eliminate pre-training requirements, improving accessibility and reproducibility of real-time DETR models.

{
    \small
    \bibliographystyle{ieeenat_fullname}
    \bibliography{main}
}

\clearpage

\appendix

{\Large \noindent \textbf{Appendix}}

\section{Implementation Details}

\label{appen:hyper}

In this section, we present the implementation details and training hyperparameters of Le-DETR. Detailed parameters are presented in Tab.~\ref{tab:hyper}. Compared with previous real-time DETR models, we have some slight modifications. First, Le-DETR-X uses 256 as the embedding dimension in the encoder, to align with the smaller dimension from our designed backbone, which is 384 in previous X-scale real-time DETR models. Also, the feedforward dimension is reduced to 1024 in the X-scale model, keeping the same as it is in the L-scale model. We use a larger total batch size to accelerate training, and they can still be trained under 12GB GPU like 8 * 2080Ti. Along with the larger total training batch size, we also use a larger base learning rate and a larger backbone learning rate. We use AdmaW as the optimizer for both the backbone pretraining and the coco training.

\begin{table*}[tbp]
    \centering
    \resizebox{1.0\textwidth}{!}{
    \begin{tabular}{lccc}
        \toprule[1pt]
        COCO Training Settings & Le-DETR-M & Le-DETR-L & Le-DETR-X\\ 
        \hdashline\noalign{\vskip 2pt}
        Backbone & EfficientNAT-M & EfficientNAT-L &  EfficientNAT-X \\
        Backbone Freezing Layer & (0, 1) &  (0, 1) & (0, 1)\\
        Embedding Dimension & 256 & 256 & 256 \\
        Feedforward Dimension & 1024 & 1024 & 1024 \\
        Encoder Layer Number & 1 & 1 & 1 \\
        NAAIFI Kernel Size & 63 & 63 & 63 \\
        Decoder Hidden Dimension & 256 & 256 & 256 \\
        Training Decoder Layer Number & 6 & 6 & 6 \\
        Inference Decoder Layer Number & 4 & 6 & 6 \\
        Queries & 300 & 300 & 300 \\
        Denoising Tokens &100 & 100 & 100 \\
        Sampling Point Number & (S: 3, M: 6, L: 3) & (S: 3, M: 6, L: 3) & (S: 3, M: 6, L: 3) \\
        Loss Function & DEIMCriterion & DEIMCriterion & DEIMCriterion  \\
        Weight of $\mathcal{L}_{\text{VFL}}$ & 1 & 1 & 1 \\
        Weight of $\mathcal{L}_{\text{BBox}}$ & 5 & 5 & 5\\
        Weight of $\mathcal{L}_{\text{GIOU}}$ & 2 &2 & 2\\
        Weight of $\mathcal{L}_{\text{FGL}}$ & 0.15 & 0.15 & 0.15\\
        Weight of $\mathcal{L}_{\text{DDF}}$ & 1.5 &1.5 & 1.5 \\
        Base Learning Rate & 1.25e-4 &1.25e-4 & 1.25e-4 \\
        Backbone Learning Rate & 5e-5 &5e-5 & 5e-5 \\
        Total Batch Size & 64 &64 & 64 \\
        Epochs & 80 & 80 & 80 \\
        EMA Decay & 0.9999 & 0.9999 & 0.9999 \\
        \midrule
        ImageNet-1K Training Settings & EfficientNAT-M & EfficientNAT-L & EfficientNAT-X \\
        \hdashline\noalign{\vskip 2pt}
        EfficientNAT Block Number & (1, 1, 2, 2) & (1, 1, 1, 4, 4) & (1, 2, 7, 15, 2) \\
        EfficientNAT Block Dimensions & (32, 64, 128, 256, 512)  & (32, 64, 128, 256, 512) & (32, 64, 128, 256, 512) \\
        Base Learning Rate & 1e-3 & 1e-3 & 1e-3 \\
        Warmup Learning Rate & 1e-6 & 1e-6 & 1e-6 \\
        Min Learning Rate & 5e-6 & 5e-6 & 5e-6 \\
        Epochs & 300 & 300 & 300\\
        Cooldown Epochs & 10 & 10 & 10 \\
        Warmup Epochs & 20 & 20 & 20\\
        Batch Size & 128 & 128 & 128 \\
        Learning Rate Scheduled & Cosine & Cosine & Cosine \\
        Weight Decay & 5e-2 & 5e-2 & 5e-2 \\
        \bottomrule[1pt]
    \end{tabular}}
    \vspace{0.5cm}
    \caption{Implementation details and training hyperparameters of our proposed method.}
    \label{tab:hyper}
\end{table*}

\section{Limitation}
\label{appen:lim}

Though we highly reduce the pre-training cost of real-time transformer-based object detection models, making it return to a low-cost process. It's necessary to claim that the training process of YOLO series models shows no need for any additional data except images of COCO train 2017. Though since our Le-DETR needs fewer epochs to converge compared with YOLO series models, so the overall training overhead is similar, we expect new transformer-based models without any pre-training. Also, the lack of neighborhood attention on ONNX and TensorRT exports may naturally hinder the practical application of our work on a small subset of cases, and we look forward to this progress.

\section{Visualization in hard scenarios}

In this section, we present visualization results demonstrating the effectiveness of our proposed Le-DETR model in handling challenging scenarios. To evaluate its robustness, we sampled a set of difficult images from the \textit{COCO Val2017} dataset. These images are particularly challenging due to the presence of one or more of the following conditions: a high density of objects within a single image, poor lighting conditions such as dim or blurred light, and motion blur, among other complexities.

\begin{figure*}[t]
    \centering
    \includegraphics[width=0.95\textwidth]{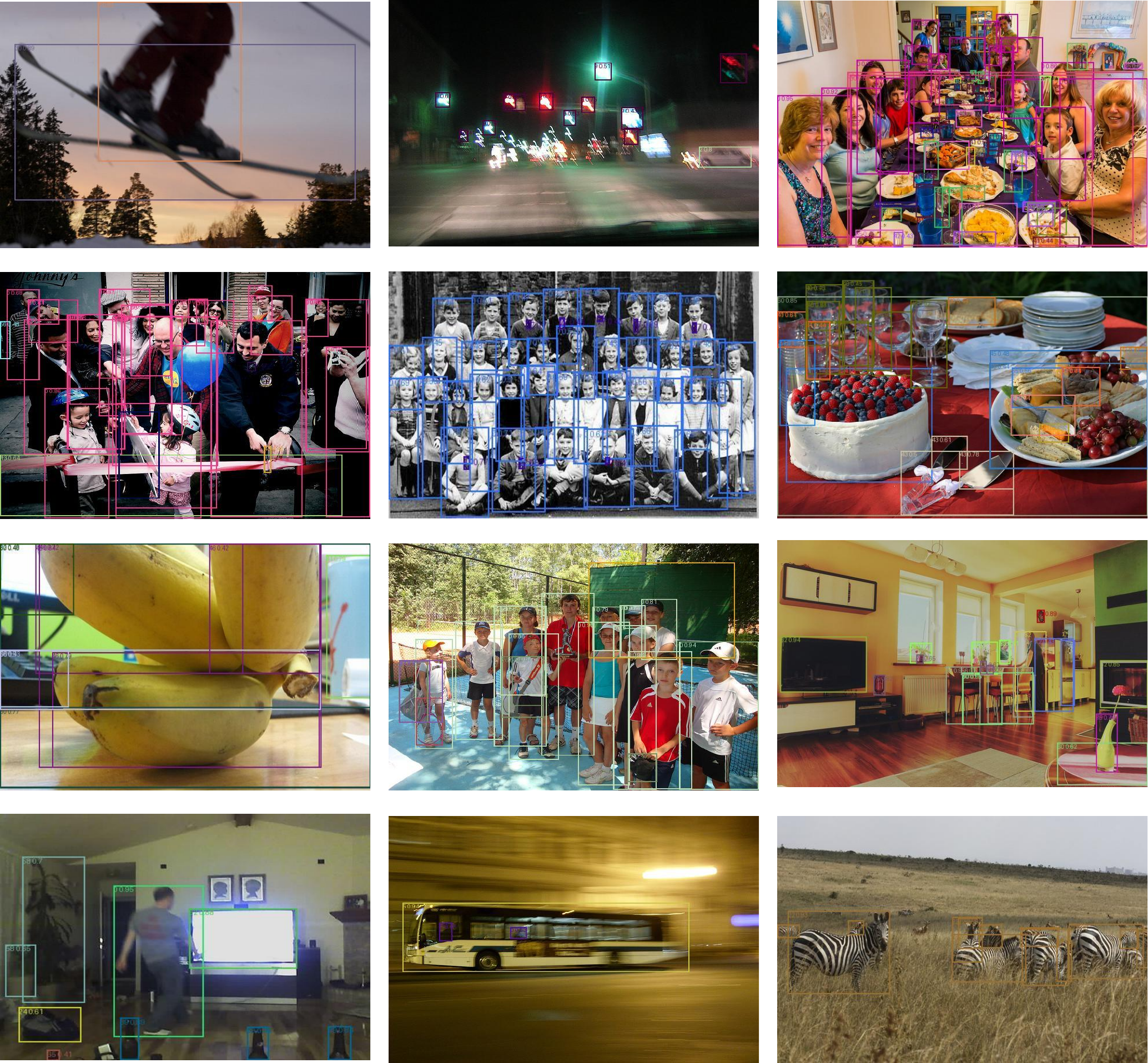}
    \caption{Visualization of applying Le-DETR-L into hard cases of object detection.}
    \label{fig:visl}
\end{figure*}
\begin{figure*}[t]
    \centering
    \includegraphics[width=0.95\textwidth]{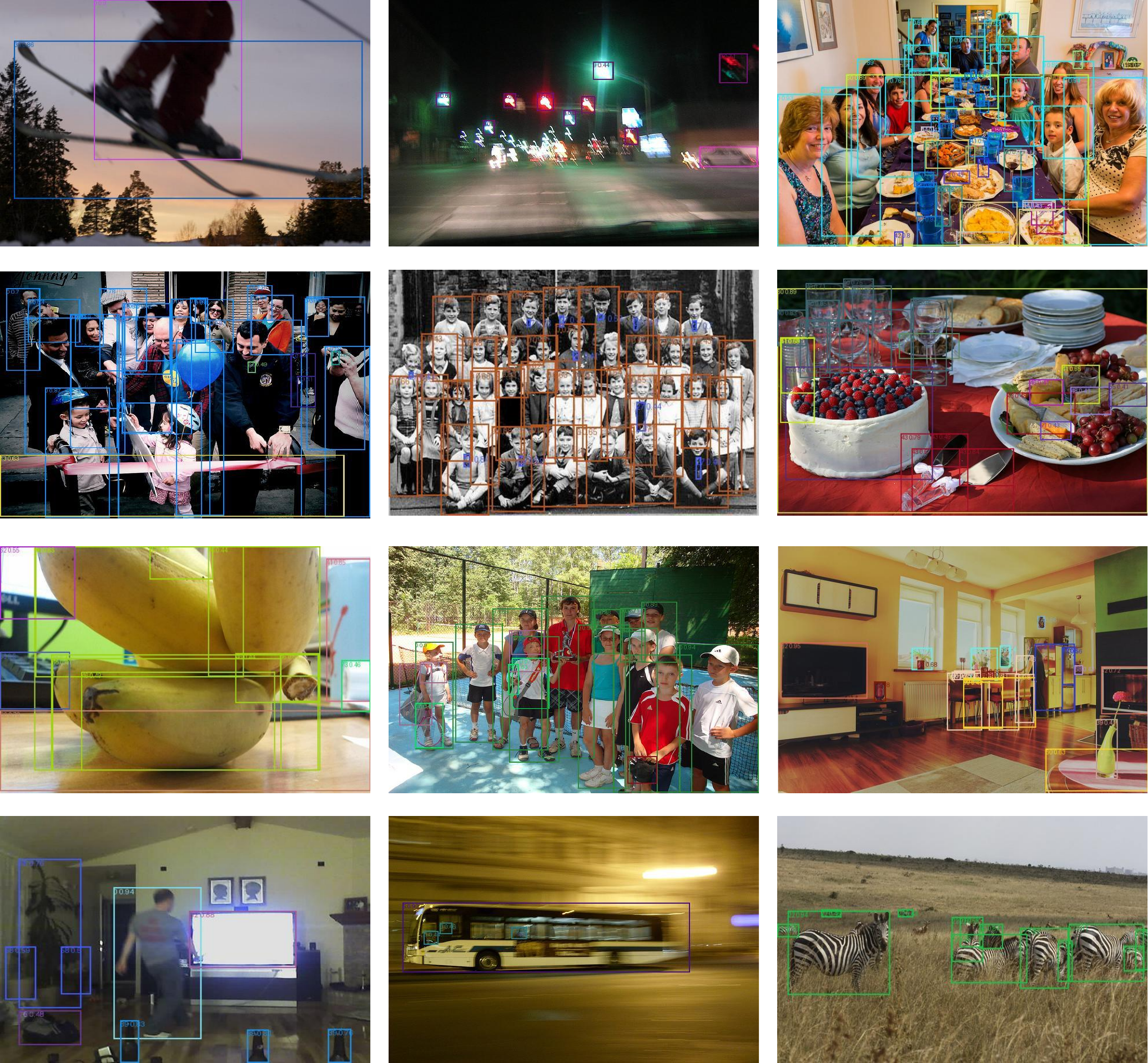}
    \caption{Visualization of applying Le-DETR-X into hard cases of object detection.}
    \label{fig:visx}
\end{figure*}

Fig. ~\ref{fig:visl} illustrates the detection results of the Le-DETR-L model, while Fig. ~\ref{fig:visx} showcases the performance of the Le-DETR-X model. The visualizations clearly indicate that our proposed model demonstrates strong capabilities in handling these hard cases. Specifically, the Le-DETR models effectively identify objects even under adverse conditions, highlighting their robustness and reliability in complex object detection tasks.

\end{document}